\documentclass[runningheads]{llncs}
\usepackage[T1]{fontenc}
\usepackage[table,xcdraw]{xcolor}
\usepackage{todonotes}
\usepackage{graphicx}
\usepackage{amsmath}
\usepackage{amssymb}
\usepackage{bm}
\usepackage{algcompatible, algorithm, algpseudocode}
\usepackage{rotating}
\usepackage{booktabs}
\usepackage{subcaption}
\usepackage{todonotes}
\usepackage{hyperref}
\usepackage{orcidlink}
\newcommand{\bx}{\mathbf{x}}
\newcommand{\by}{\mathbf{y}}

\newcommand{\bgamma}{\bm{\gamma}}

\newcommand{\btheta}{\bm{\theta}}
\newcommand{\fvec}{\mathbf{f}}
\newcommand{\mX}{\mathcal{X}}
\newcommand{\mP}{\mathcal{P}}

\newcommand{\mF}{\mathcal{F}}

\newcommand{\Real}{\mathbb{R}}
\newcommand{\Exp}{\mathbb{E}}

\begin{document}
\title{Accelerating Multi‑Objective Bayesian Optimisation via Predictive‑Gradient Catalysts}
\titlerunning{Accelerating MOBO via Predictive‑Gradient Catalysts}
% \author{Alma Rahat\inst{1}\orcidID{0000-1111-2222-3333} \and

\author{Alma Rahat\thanks{Corresponding author}\inst{1}\orcidlink{0000-0002-5023-1371} \and
Tinkle Chugh\inst{2}\orcidlink{0000-0001-5123-8148} \and
Jonathan Fieldsend\inst{2}\orcidlink{0000-0002-0683-2583}
\and Richard~Allmendinger\inst{3}\orcidlink{0000-0003-1236-3143}}

\authorrunning{Rahat et al.}

\institute{Loughborough University, Loughborough, LE11 3TU, U.K. \\
\email{A.Rahat@lboro.ac.uk}\and
University of Exeter, Exeter EX4 4QD, U.K.
\\ \email{\{T.Chugh,J.E.Fieldsend\}@exeter.ac.uk} \and
The University of Manchester, Manchester M15 6PB, U.K.\\\email{richard.allmendinger@manchester.ac.uk}}
% %
\maketitle              % typeset the header of the contribution
\begin{abstract}
This paper presents a general acceleration mechanism for multi-objective Bayesian optimisation (MOBO) that leverages Gaussian process predictive gradients as auxiliary signals. Rather than replacing existing Pareto-compliant acquisition functions, the proposed approach augments them with local stationarity information derived from surrogate-derived gradients, enabling faster convergence toward the global Pareto set under limited evaluation budgets. Two catalyst instantiations are investigated: an adaptive Multiple-Gradient Descent Algorithm-Based Catalyst (MGDA) and a predefined-weight variant that enables focused exploration when budgets are tight. Experiments on the DTLZ benchmark suite (using 2 objectives and 10 decision variables) show that predictive gradient catalysis can deliver significant acceleration compared to other acquisition functions (EHVI, AugTch, tMPoI, SAF) when surrogates are accurate, particularly for stationary problems.
% \footnote{Supplementary material is available at \href{https://doi.org/10.5281/zenodo.19649281}{Zenodo: 19649281}; code and data will follow.}
\footnote{Supplementary material, code, and data are available at \href{https://doi.org/10.5281/zenodo.19649281}{Zenodo: 19649281}.}

\end{abstract}
%
%
%
% \section{Introduction}

% To the best of our knowledge, no existing multi-objective Bayesian optimisation method exploits GP predictive gradients as auxiliary, catalyst‑like signals to accelerate the optimisation of acquisition functions or the navigation of the search space.

\section{Introduction}
Bayesian optimisation (BO) is a widely used framework for optimising expensive, noisy, or black‑box objective functions. By combining probabilistic surrogate models, most commonly Gaussian processes, with acquisition functions that balance exploration and exploitation, BO can identify high-quality solutions using relatively few expensive evaluations. This has led to successful applications in machine learning, engineering design, and simulation‑based optimisation~\cite{shahriari2015taking,garnett2023bayesian}.

Recent work has explored the use of gradient information to accelerate BO. Early approaches focus on incorporating \emph{observed} gradients, obtained via adjoint methods or automatic differentiation, into surrogate models and acquisition functions. In particular, Wu et al.~\cite{wu2017bayesian} showed that using exact gradients within a one‑step look-ahead knowledge‑gradient framework can substantially reduce sample complexity. More recently, Perrin and Le~Riche proposed an acceleration strategy that exploits \emph{predictive} gradients of Gaussian process surrogates to bias acquisition toward locally optimal regions~\cite{perrin2024bayesian}. Although effective, this approach is limited to single-objective settings and is tightly coupled to a specific acquisition function.

However, many real-world problems are inherently multi-objective, requiring the optimisation of several competing criteria and yielding a set of trade‑off solutions rather than a single optimum. In such problems, small design changes can significantly shift how performance is distributed across objectives, making efficient exploration of the Pareto set particularly challenging when evaluations are expensive. Although gradient-based stationarity notions play a central role in deterministic multi‑objective optimisation, their use within multi-objective Bayesian optimisation (MOBO) has remained largely unexplored.

This paper addresses this gap by introducing a general acceleration mechanism for MOBO that leverages Gaussian process \emph{predictive gradients} as auxiliary, catalyst-like signals. Rather than replacing existing acquisition functions, the proposed framework augments them with local stationarity information derived from surrogate gradients, enabling faster convergence toward the Pareto set under limited evaluation budgets.

% \paragraph{Contributions.}
The main \textit{contributions} of this paper are:
\begin{itemize}
  \item A catalytic framework for MOBO that exploits GP predictive gradients to accelerate convergence without altering underlying acquisition functions.
  \item Two catalyst instantiations grounded in Pareto stationarity theory: an adaptive  Multiple-Gradient Descent Algorithm (MGDA)-based strategy and a predefined-weight approach for focused trade-off exploration.
  \item A principled integration of catalytic signals with standard Pareto-compliant acquisition functions via augmented Tchebycheff scalarisation.
  \item An extensive evaluation on the DTLZ benchmark suite showing consistent acceleration for (approximately) stationary problems.
  
  % An extensive evaluation on the DTLZ benchmark suite showing consistent acceleration when surrogate models are accurate.
  % and approximately stationary.
\end{itemize}

The remainder of the paper is organised as follows. Section~\ref{sec:background} reviews background material on multi-objective optimisation, Gaussian processes, and MOBO. Section~\ref{sec:cat_framework} presents the proposed catalytic framework. Experimental settings and results are discussed in Sections~\ref{sec:exp_settings} and~\ref{sec:results}, followed by conclusions in Section~\ref{sec:concl}.

\section{Background}
\label{sec:background}
We now set out the background material underpinning this work.

\subsection{Multi-Objective Optimisation and Pareto Stationarity}

Let $\bx \in \Real^n$ be an $n$-dimensional decision vector. Without loss of generality, an unconstrained multi-objective optimisation problem can be defined as
\begin{align}
    \min_{\bx \in \mX} \, \fvec(\bx),
\end{align}
where $\fvec(\bx) = (f_1(\bx), \dots, f_M(\bx))^\top$ is an $M$-dimensional objective vector, $f_i(\cdot)$ denotes the $i$th objective, and $\mX \subseteq \Real^n$ is the feasible space.

Solving a multi-objective problem yields a set of optimal trade-off solutions rather than a single optimal point. A solution is considered desirable if no other feasible solution can improve one objective without worsening at least one other. This leads to the standard concept of Pareto optimality. A vector $\bx' \in \mX$ is Pareto optimal if there exists no other vector $\bx'' \in \mX$ that dominates it; that is,
\begin{itemize}
    \item $\bx''$ is at least as good as $\bx'$ in all objectives, and
    \item $\bx''$ is strictly better in at least one objective.
\end{itemize}

Formally, the Pareto optimal set $\mP^*$ is defined as
\begin{align}
    \mP^*
    = \Big\{
        \bx' \in \mX ~\Big|~
        &\nexists \bx'' \in \mX :
        \left(
            \forall i \in \{1,\dots,M\},\ f_i(\bx'') \le f_i(\bx')
        \right)
        \notag\\
        &\qquad\qquad\qquad\land~
        \left(
            \exists j \in \{1,\ldots,M\},\ f_j(\bx'') < f_j(\bx')
        \right)
    \Big\}.
\end{align}

The Pareto front $\mF^*$ is the image of the Pareto set in the objective space, i.e.,
$
    \mF^* = \left\{ \mathbf{f}(\bx) : \bx \in \mP^* \right\}.
$
 
The necessary conditions for Pareto stationarity-meaning that any sufficiently small perturbation in the decision space around a stationary point $\bx$ yields an equivalent or dominated solution-are given for unconstrained multi-objective optimisation problems in \cite{miettinen1999nonlinear} as:
\begin{align}
\label{eq:FJ_conditions}
    \bgamma \cdot \nabla \fvec(\bx)
    = \sum_{i=1}^M \gamma_i \nabla f_i(\bx)
    = 0,
\end{align}
where $\bgamma = (\gamma_1, \dots, \gamma_M)^\top$ denotes a vector of convex combination coefficients with $\gamma_i \in [0,1]$ and $\sum_i \gamma_i = 1$.
These conditions, commonly referred to as the Fritz--John (FJ) conditions, imply that at a Pareto stationary point, either all objectives are simultaneously minimised or their improvement directions are in conflict such that the weighted sum of their gradients cancels out, leaving no feasible descent direction that improves all objectives simultaneously.

Importantly, the FJ conditions are \emph{necessary but not sufficient} for global Pareto optimality. Sufficiency holds only under additional assumptions; see, for instance, Censor~\cite{censor1977pareto}. Consequently, the condition may also be satisfied at points that are locally efficient yet globally dominated. Hence, the FJ condition alone guarantees only \emph{local} Pareto efficiency rather than global optimality.

% \todo[inline]{For Fritz-John conditions for Pareto optimality, see \cite{miettinen1999nonlinear}, chapter 3. }

\subsection{Gaussian Process Regression}

Gaussian Process (GP) models provide a flexible Bayesian framework for regression, offering closed-form posterior predictive distributions that quantify both expected objective values and the associated epistemic uncertainty. The predictive mean represents the model's current best estimate of the objective, while the predictive variance reflects the local scarcity of observations. This dual representation naturally underpins the exploration--exploitation trade-off, making GPs a powerful choice for surrogate-assisted optimisation \cite{de2021greed}.

For each objective $i$, we consider a dataset $\mathcal{D}_i = \{(\bx^t, y^t = f_i(\bx^t))\}_{t=1}^{T}$ of size $T$. A GP equipped with hyperparameters $\btheta_i^*$ yields the Gaussian predictive distribution
\begin{align}
    \hat{f}_i(\bx) \sim p(f_i(\bx)\mid\mathcal{D}_i, \btheta_i^*)
    =
    \mathcal{N}\!\left(
        \mu_i(\bx),\,
        \sigma_i^2(\bx)
    \right).
\end{align}
The predictive moments \cite{rasmussen2006gaussian} are given by
\begin{align}
    \mu_i(\bx) &= \kappa(\bx, X) K^{-1}\by_i,\\
    \sigma_i^2(\bx) &= \kappa(\bx,\bx)
    - \kappa(\bx,X) K^{-1}\kappa(X,\bx),
\end{align}
where $X \in \mathbb{R}^{T\times n}$ contains the input locations, $\by_i \in \mathbb{R}^{T}$ the corresponding objective evaluations, and $K$ is the covariance matrix induced by the kernel $\kappa(\cdot,\cdot\,|\,\btheta_i^*)$. The vector $\kappa(\bx,X)$ denotes the covariances between the query point $\bx$ and all observations.

The kernel plays a central role in encoding smoothness and structural assumptions about the underlying objective. In this work, we employ the Matérn-$5/2$ kernel, which has been recommended for real-world optimisation tasks \cite{snoek2012practical}. Hyperparameters $\btheta_i^*$ are inferred via maximum likelihood using the L-BFGS optimiser \cite{liu1989limited} with ten random restarts; see \cite{gpy2014} for details. %{\color{red} JF: was it L-BFGS, or L-BFGS-B?}

Because the predictive distribution is Gaussian, the expected gradient of $f_i$ is simply the gradient of its predictive mean. Differentiating yields~\cite{solak2003derivative}:
\begin{align}
    \Exp\left[\nabla \hat{f}_i(\bx)\right]
    =
    \left(
        \frac{\partial \mu_i(\bx)}{\partial x_j}
        \;\bigg|\;
        j = 1,\dots,n
    \right)^\top
    =
    \left(
        \frac{\partial \kappa(\bx, X)}{\partial x_j} K^{-1}\by_i
        \;\bigg|\;
        j = 1,\dots,n
    \right)^\top.
\end{align}

\begin{figure}[t!]
    \centering
    \includegraphics[width=0.9\linewidth, trim={5 20 5 5mm}, clip = true]{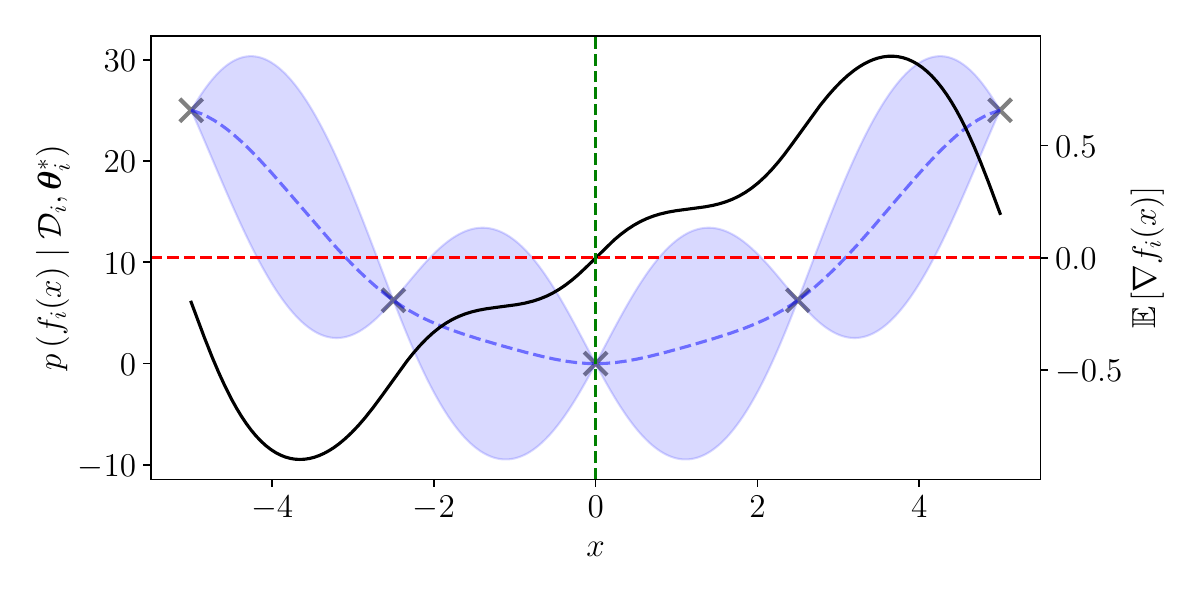}
    \caption{
        Illustration of the predictive mean, uncertainty, and expected gradient of a GP model trained on the Sphere function $f_i(x) = x^2$. 
        The predictive mean (blue dashed line) and uncertainty band (light blue) are shown alongside the expected gradient (solid black line), which correctly vanishes at the true minimum (green dashed line). The horizontal red dashed line indicates the zero‑gradient target, $\Exp[\nabla f_i(\bx)]=0$.
        Training data are indicated by grey crosses. 
    }
    \label{fig:GP-der}
\end{figure}

Figure~\ref{fig:GP-der} visualises these quantities for a GP model of the Sphere function, illustrating how the GP predictive gradient correctly identifies the stationary point at the true optimum with a small dataset of size five.

Although we assume noise-free observations in this example, GP regression readily accommodates noisy data through an additive Gaussian noise term. For instance, with homoscedastic noise of variance $\sigma_{noise}^2$, the predictive variance becomes $\sigma_i^2(\bx) + \sigma_{noise}^2$.

A standard strategy for constructing multi-objective surrogate models is to assume conditional independence between objectives and to model each objective with an individual GP. Under this assumption, the joint predictive model factorises across objectives, yielding:
\begin{align}
    \hat{\fvec}(\bx)
    \mid \bm{\mathcal{D}}
    \;\sim\;
    \prod_{i=1}^{M}
    p\!\left(f_i(\bx)\right)
    =
    \prod_{i=1}^{M}
    \mathcal{N}\!\left(
        \mu_i(\bx),\,
        \sigma_i^2(\bx)
    \right),
\end{align}
where $\bm{\mathcal{D}} = \{X, Y\}$ denotes the training dataset, consisting of the input matrix $X$ and the corresponding objective evaluation matrix $Y = (\by_1, \dots, \by_M)^\top$. 

We adopt this independent multi-surrogate formulation as the foundation of our Bayesian multi-objective optimisation framework. This choice is supported by prior studies showing that independent per-objective GPs generally outperform mono-surrogate approaches, in which a scalarised version of the multi-objective problem is modelled using a single GP~\cite{rahat2017alternative,chugh2022mono,Chugh2019ScalarizingOptimization}. 

\subsection{Multi-Objective Bayesian Optimisation}

Multi-objective Bayesian Optimisation (MOBO) has seen significant growth in the past two decades, with early influential contributions appearing around 2006~\cite{emmerich2006single,knowles2006parego,keane2006statistical}. This rise has been driven by the need to efficiently approximate the Pareto set of expensive black-box problems. 

The general MOBO workflow is conceptually straightforward: an initial dataset is generated using a space-filling design such as Latin Hypercube Sampling~\cite{mckay1992latin}, producing a set of candidate solutions $X$ and their corresponding objective evaluations $Y$. A mono or multi-surrogate GP model is then trained on this dataset, yielding a multivariate predictive distribution over the objectives. An acquisition function is employed to select the next candidate solution expected to improve the current approximation of the Pareto set, trading off exploitation (via the predictive mean) and exploration (via predictive uncertainty). The selected solution is evaluated using the true expensive objectives, added to the dataset, and the surrogate is retrained. This iterative procedure continues until the evaluation budget is exhausted. A high-level outline of the method is provided in the  \href{https://doi.org/10.5281/zenodo.19649281}{supplementary material}.

Acquisition functions in this setting are typically constructed from a Pareto-compliant utility indicator, meaning that any newly identified non-dominated solution yields a strictly better indicator value relative to the current approximation of the Pareto set $\tilde{\mP}$. 
Given such an indicator, the standard BO approach computes its expected improvement by integrating the utility over the predictive distribution in the objective space, unless the function already accounts for the predictive distribution.
The aim of such approaches is to yield an acquisition value that balances both the predicted objective performance and the associated epistemic uncertainty.

% In the following, we summarise the widely used acquisition functions that serve as baselines in our empirical study.
In this study, we employ a set of widely used acquisition functions-namely, Expected Hypervolume Improvement (EHVI) \cite{emmerich2006single}, Augmented Tchebycheff decomposition (AugTch) \cite{chugh2022mono}, the Transformed Minimum Probability of Dominance (tMPoI) \cite{rahat2017alternative}, and the Summary Attainment Front distance (SAF) \cite{bautista2009sequential,svenson2016multiobjective} -- as baseline acquisition functions for acceleration. Detailed descriptions of these acquisition functions are provided in the \href{https://doi.org/10.5281/zenodo.19649281}{supplementary material}.

% \subsubsection{Quasi Monte Carlo Approximation} - no space for this; may be in the appendix

\section{Proposed Catalytic Framework}
\label{sec:cat_framework}

The overarching goal of this work is to identify the global Pareto set as efficiently as possible under a limited budget of expensive function evaluations. In particular, we investigate whether predictive gradients obtained from multivariate GP surrogate models can be used to enhance existing Pareto-compliant acquisition functions by acting as a local catalyst that focuses the search.

According to the FJ optimality conditions for Pareto stationarity (see Equation~\eqref{eq:FJ_conditions}), points in the decision space where a suitable convex combination of objective gradients vanishes correspond to locally or globally Pareto‑stationary solutions. When employing a multivariate GP surrogate, these conditions can be approximated in expectation through the predictive gradients of the surrogate model,
\begin{align}
\boldsymbol{\gamma} \cdot \nabla \hat{\fvec}(\bx)
=
\boldsymbol{\gamma} \cdot
\bigl(
\mathbb{E}[\nabla \hat f_i(\bx)]
\;\big|\;
i=1,\dots,M
\bigr).
\end{align}

Since the FJ conditions are necessary but not sufficient for Pareto optimality, they do not distinguish between local and global Pareto sets. Nevertheless, vanishing predictive gradients are highly informative within local neighbourhoods of the decision space. This motivates their use as a \emph{catalyst} for Pareto‑compliant acquisition functions: locally inefficient points can be filtered out, while solutions satisfying
\(
\lVert
\boldsymbol{\gamma} \cdot \nabla \hat{\fvec}(\bx^\star)
\rVert = 0
\)
are identified as locally Pareto‑efficient.

By contrast, Pareto‑compliant acquisition functions, given sufficiently accurate surrogate models, can discriminate between local and global Pareto sets. Our framework therefore, combines both criteria: the acquisition function provides global guidance, while predictive-gradient information focuses the search locally.

We formalise this as a bi‑objective decision problem,
\begin{align}
\max \; g_1(\bx) &\equiv \max \; \alpha(\bx), \\
\min \; g_2(\bx) &\equiv \min \;
\lVert
\boldsymbol{\gamma} \cdot \nabla \hat{\fvec}(\bx)
\rVert,
\end{align}
where \(\alpha(\bx)\) denotes a Pareto‑compliant acquisition function. Assuming equal importance of global and local criteria, we combine them using an augmented Tchebycheff scalarisation~\cite{knowles2006parego},
\begin{align}
\alpha_{\mathrm{CAT}}(\bx)
=
\max\!\left(
\frac{g_1(\bx)}{2},
-
\frac{g_2(\bx)}{2}
\right)
-
\rho\!\left(
\frac{g_1(\bx)}{2}
-
\frac{g_2(\bx)}{2}
\right),
\label{eq:tch_inside}
\end{align}
where the negation converts the minimisation of \(g_2\) into a maximisation problem. This formulation penalises candidate solutions that are inferior with respect to either global or local considerations---$\rho$ is a tiny positive penalty parameter which effectively enables discrimination between dominating and equivalent points under the two objectives. An adaptation of weights between $g_1$ and $g_2$ can be explored in the future.

\subsection{Multiple‑Gradient Descent Algorithm-Based Catalyst (MGDA)}

In the first strategy, the weight vector \(\boldsymbol{\gamma}\) is determined adaptively using the Multiple‑Gradient Descent Algorithm (MGDA)~\cite{schaffler2002stochastic,desideri2012mutiple}. MGDA identifies, at a given point \(\bx\), the convex combination of objective gradients with minimum Euclidean norm,
\begin{align}
\boldsymbol{\gamma}^\star
=
\arg\min_{\gamma_i \ge 0,\; \sum_i \gamma_i = 1}
\left\lVert
\sum_{i=1}^{M}
\gamma_i \nabla f_i(\bx)
\right\rVert^2.
\end{align}

This optimisation admits a closed‑form solution,
\(
\boldsymbol{\gamma}^\star
=
(\mathbf{H}^{-1}\mathbf{1})/
(\mathbf{1}^\top \mathbf{H}^{-1}\mathbf{1}),
\)
where \(\mathbf{H} = \mathbf{G}^\top \mathbf{G}\) is the Gram matrix of the Jacobian
\(\mathbf{G} := \nabla \fvec(\bx)\).
The resulting direction satisfies the FJ stationarity conditions and adapts optimally to the local objective geometry.

\subsection{Predefined Weight Vectors (W)}

% In the second strategy, we consider a predefined and finite set of weight vectors that discretise the normalised objective space $\fvec(\bx)$ following the method~\cite{Das1998} (see \href{https://doi.org/10.5281/zenodo.XXXXXXX}{supplementary material}).
% of Knowles~\cite{knowles2006parego}. \todo[inline]{AR: add description in supplementary.}
In the second strategy, we discretise the objective space $\fvec(\bx)$ using predefined weight vectors as in~\cite{Das1998}, after applying min-max normalisation computed from the observed objective values~$Y$.
This approach restricts the search to a coarse set of directions and can improve robustness and focus under a limited evaluation budget.

A limitation of this strategy is that predefined weight vectors are not guaranteed to satisfy the FJ conditions, as the chosen directions may not align with the locally optimal convex combination of gradients. Nevertheless, for expensive optimisation problems, this restriction can be advantageous by prioritising targeted exploration of promising trade-off regions rather than exploring all directions simultaneously.

\begin{figure*}[t!]
    \centering
    % --- Row 1: MGDA ---
    \begin{subfigure}[t]{0.48\textwidth}
        \centering
        \includegraphics[width=\linewidth, trim={10 10 10 6mm}, clip=true]{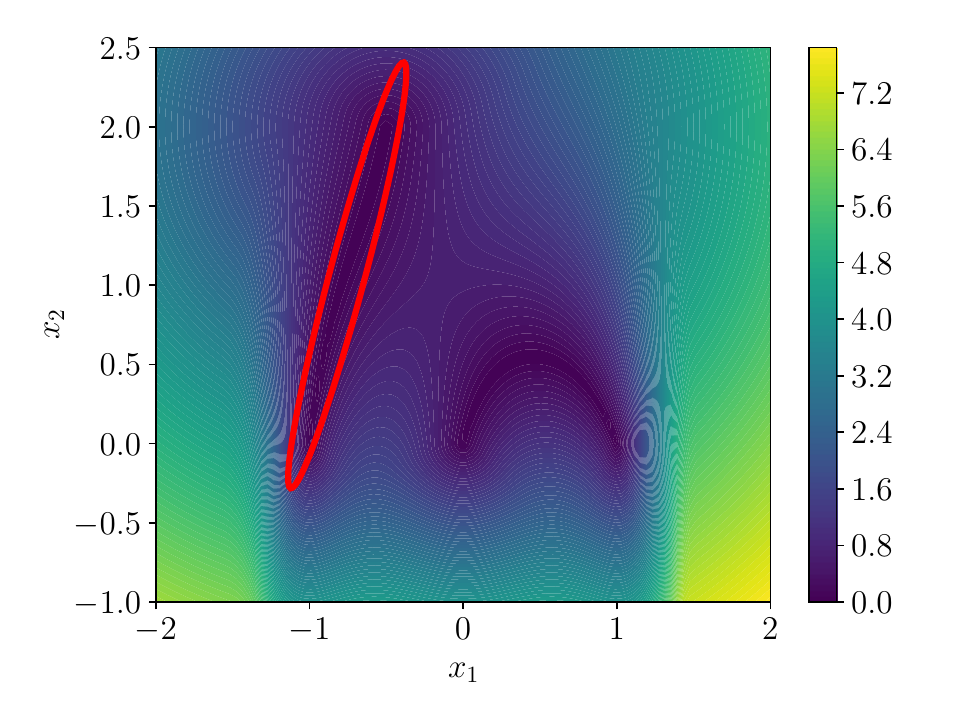}
        \caption{MGDA (true landscape)}
        \label{fig:mgda-true}
    \end{subfigure}
    \hfill
    \begin{subfigure}[t]{0.48\textwidth}
        \centering
        \includegraphics[width=\linewidth, trim={10 10 10 6mm}, clip=true]{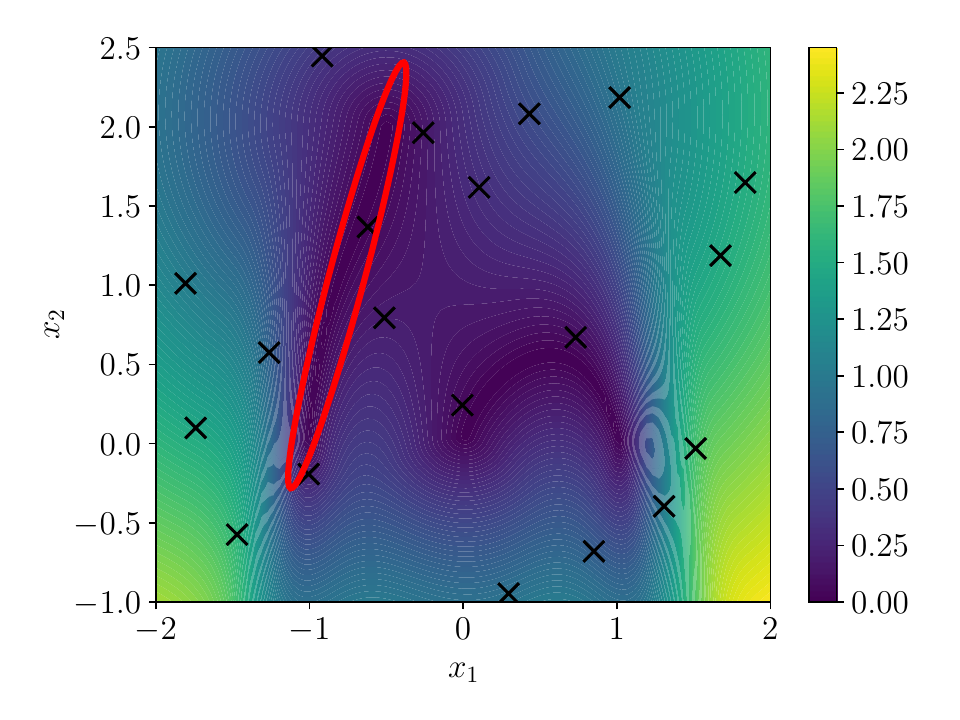}
        \caption{MGDA (GP approximation)}
        \label{fig:mgda-gp}
    \end{subfigure}

    \vspace{0.6em}

    % --- Row 2: Predefined weights (W) ---
    \begin{subfigure}[t]{0.48\textwidth}
        \centering
        \includegraphics[width=\linewidth, trim={10 10 10 6mm}, clip=true]{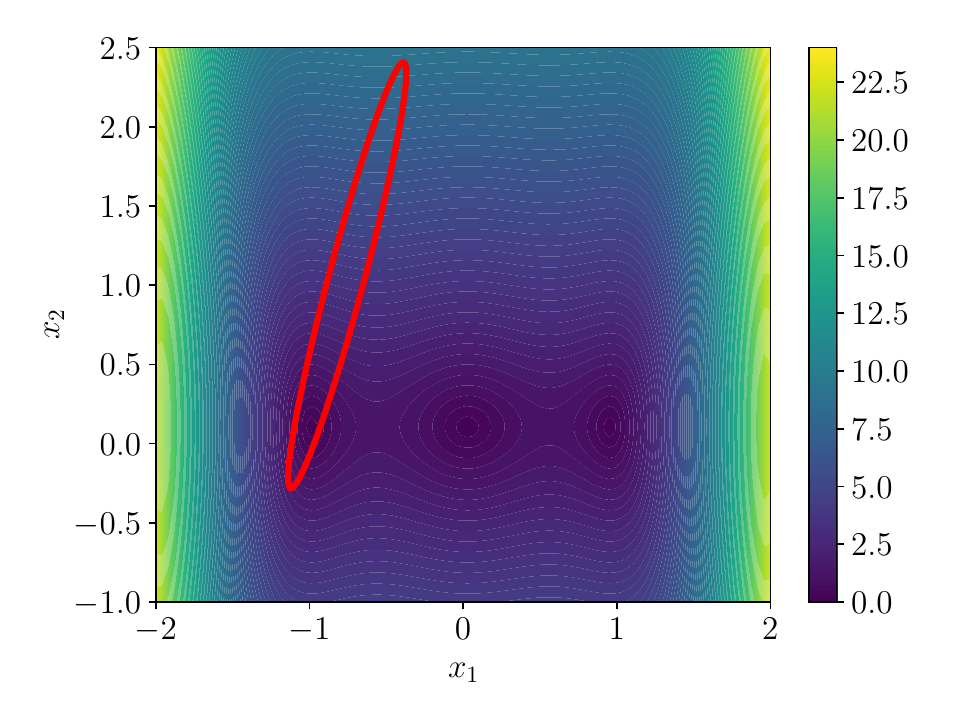}
        \caption{Predefined weights $\boldsymbol{\gamma}=(0.9,0.1)^\top$ (true landscape)}
        \label{fig:w-true}
    \end{subfigure}
    \hfill
    \begin{subfigure}[t]{0.48\textwidth}
        \centering
        \includegraphics[width=\linewidth, trim={10 10 10 6mm}, clip=true]{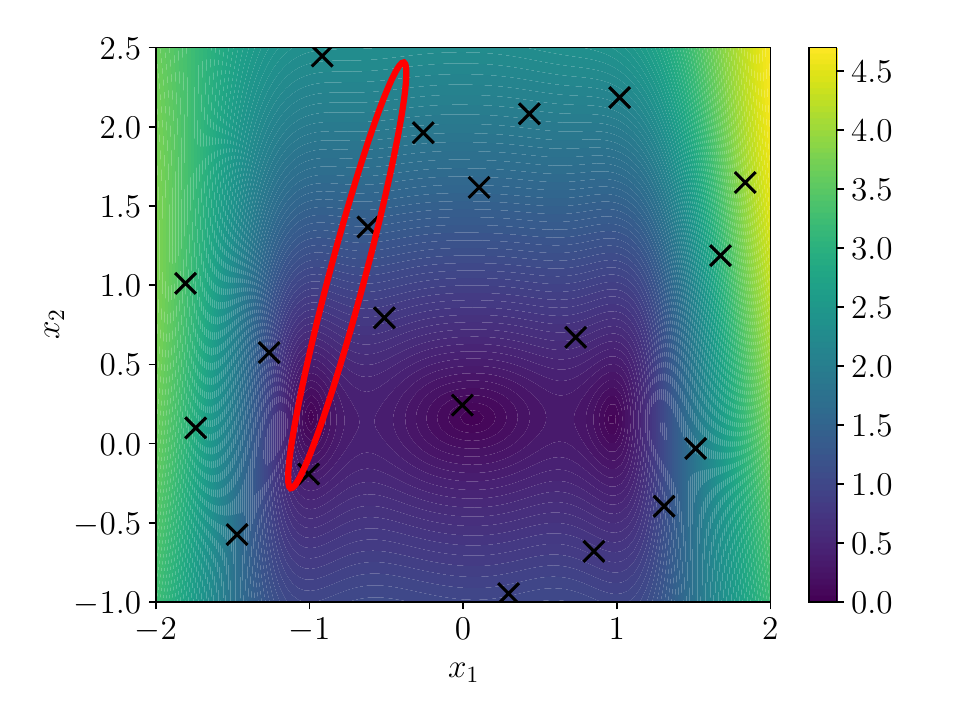}
        \caption{Predefined weights $\boldsymbol{\gamma}=(0.9,0.1)^\top$ (GP approximation)}
        \label{fig:w-gp}
    \end{subfigure}

    \caption{
    Gradient‑magnitude landscapes for MGDA (top) and a predefined weight vector $\boldsymbol{\gamma}=(0.9,0.1)^\top$ (bottom) on the bi‑objective problem
$\fvec(\bx)=\bigl(x_1^4-2x_1^2+2x_2^2+1,\,(x_1+0.5)^2+(x_2-2)^2\bigr)$.
Red ellipses mark the global Pareto set, while (near‑)zero regions (purple) outside this set indicate local Pareto‑stationary regions, which are not guaranteed to satisfy FJ conditions for predefined weights.
MGDA captures stationarity along a complete Pareto set, whereas the predefined weighting highlights a specific trade‑off region.
GP surrogates trained with only 20 initial samples (black crosses) closely approximate the true landscapes.
%     Landscapes induced by gradient magnitudes for MGDA (top) and a predefined weight vector $\boldsymbol{\gamma}=(0.9,0.1)^\top$ (bottom) on the bi‑objective problem
% \(\fvec(x_1,x_2)=\bigl(x_1^4-2x_1^2+2x_2^2+1,\,(x_1+0.5)^2+(x_2-2)^2\bigr)\).
% Red ellipses indicate the location of the global Pareto set, whereas regions exhibiting (near‑)zero values (shown in purple) outside this set correspond to local Pareto‑stationary regions; for predefined weights, such regions are not guaranteed to satisfy FJ stationarity conditions.
% MGDA captures Pareto stationarity across the full front, while the predefined weighting emphasises a specific trade‑off region.
% In all cases, Gaussian process surrogates trained using only 20 initial design points (black crosses in the right panels) closely approximate the true landscapes.
}
    \label{fig:mgda-w-2x2}
\end{figure*}

Figure~\ref{fig:mgda-w-2x2} illustrates the function landscapes induced by the proposed catalytic strategies for an illustrative bi‑objective problem.

% \todo{We used $w$ for weight vectors, can you use $w$ instead of $\gamma$ for c and d in Figure 2? We can change it later if its take too much of time.}
\subsection{Limitations}

The proposed catalytic framework relies on the accuracy of surrogate predictive gradients, which may be unreliable in regions with sparse data or poor model fit. Moreover, the MGDA-based variant can dilute the search effort under strict evaluation budgets, while the predefined weight approach trades FJ stationarity guarantees for increased robustness. Despite these limitations, combining local predictive-gradient information with global Pareto-compliant acquisition functions offers a practical balance between exploration and exploitation in expensive multi‑objective optimisation, particularly as predictive gradients are smooth and enable additional neighbourhood‑level exploitation.

\section{Experimental Settings}
\label{sec:exp_settings}

We evaluate all proposed methods on the DTLZ benchmark suite~\cite{Deb2005}, considering all seven problems (DTLZ1--DTLZ7) with two objectives and a ten-dimensional decision space. These problems are widely used to assess convergence, diversity, scalability, and robustness in multi-objective optimisation.

Each run starts from an initial design of size \(I = 10n\) and is performed under a total evaluation budget of \(B = 30n\). For each problem--method pair, we conduct 11 independent optimisation runs. To enable paired statistical comparisons, all methods share identical initial designs within each run, ensuring that observed performance differences are attributable to the optimisation strategy rather than stochastic sampling effects.

Acquisition functions are optimised using Bi-POP CMA-ES \cite{hansen2009benchmarking} with a budget of $5000n$ evaluations; W uses $10M$ predefined weight vectors using ~\cite{Das1998}. For AugTch, two parameters are set as $|\bm{\lambda}| = 10M = 20$ and $\rho = 0.05$. Performance is measured using the final hypervolume indicator with respect to predefined reference points; further details are provided in the \href{https://doi.org/10.5281/zenodo.19649281}{supplementary material}. Statistical significance is assessed via one-tailed Wilcoxon signed-rank tests evaluating directional improvement across paired runs, with Bonferroni correction yielding a corrected significance level of $\alpha = 0.05/4$.

To assess surrogate model quality independently of optimisation outcomes, we apply Pareto-front-only leave-one-out cross-validation (PF-LOOCV). In MO\-BO, the key utility of a surrogate is its ability to identify potentially non-dominated points---predictive performance elsewhere is not necessarily reflective of optimisation quality. Accuracy is quantified using the standardised log loss (SLL)~\cite{rasmussen2006gaussian}, defined relative to a naïve Gaussian baseline predicting the empirical mean. An SLL of zero corresponds to baseline performance, negative values indicate improvement, and positive values indicate degradation. We report the median of the median SLL across validation folds together with the median absolute deviation (MAD) across runs, summarising both accuracy and variability in surrogate performance.

\section{Results and Analysis}
\label{sec:results}

We begin by analysing the pairwise significance map shown in Figure~\ref{fig:traffic-light}, which summarises the effect of acceleration strategies across all problems. The experiments cover the seven DTLZ problems and four baseline acquisition functions: EHVI, AugTch, tMPoI, and SAF. Two classes of acceleration mechanisms (referred to as \emph{catalysts}) are considered: MGDA-based acceleration and predefined weighted-gradient acceleration (W), yielding a total of eight algorithmic variants.

\begin{figure}[t!]
    \centering
{\fontfamily{lmtt}\scriptsize 

\begin{tabular}{lccc||ccc||ccc||ccc|}
              & \multicolumn{3}{c}{EHVI}                                                                                                                & \multicolumn{3}{c}{AugTch}                                                                                                              & \multicolumn{3}{c}{tMPol}                                                                                                               & \multicolumn{3}{c}{SAF}                                                                                                                 \\
DTLZ1& \cellcolor[HTML]{C0C0C0}B$\rightarrow$MGDA & \cellcolor[HTML]{C0C0C0}B$\rightarrow$W & \cellcolor[HTML]{C0C0C0}MGDA$\rightarrow$W & \cellcolor[HTML]{C0C0C0}B$\rightarrow$MGDA & \cellcolor[HTML]{C0C0C0}B$\rightarrow$W & \cellcolor[HTML]{C0C0C0}MGDA$\rightarrow$W & \cellcolor[HTML]{C0C0C0}B$\rightarrow$MGDA & \cellcolor[HTML]{C0C0C0}B$\rightarrow$W & \cellcolor[HTML]{C0C0C0}MGDA$\rightarrow$W & \cellcolor[HTML]{C0C0C0}B$\rightarrow$MGDA & \cellcolor[HTML]{C0C0C0}B$\rightarrow$W & \cellcolor[HTML]{C0C0C0}MGDA$\rightarrow$W \\
DTLZ2& \cellcolor[HTML]{32CB00}B$\rightarrow$MGDA & \cellcolor[HTML]{32CB00}B$\rightarrow$W & \cellcolor[HTML]{32CB00}MGDA$\rightarrow$W & \cellcolor[HTML]{32CB00}B$\rightarrow$MGDA & \cellcolor[HTML]{32CB00}B$\rightarrow$W & \cellcolor[HTML]{32CB00}MGDA$\rightarrow$W & \cellcolor[HTML]{32CB00}B$\rightarrow$MGDA & \cellcolor[HTML]{32CB00}B$\rightarrow$W & \cellcolor[HTML]{FE0000}MGDA$\rightarrow$W & \cellcolor[HTML]{C0C0C0}B$\rightarrow$MGDA & \cellcolor[HTML]{32CB00}B$\rightarrow$W & \cellcolor[HTML]{32CB00}MGDA$\rightarrow$W \\
DTLZ3& \cellcolor[HTML]{C0C0C0}B$\rightarrow$MGDA & \cellcolor[HTML]{C0C0C0}B$\rightarrow$W & \cellcolor[HTML]{C0C0C0}MGDA$\rightarrow$W & \cellcolor[HTML]{FE0000}B$\rightarrow$MGDA & \cellcolor[HTML]{C0C0C0}B$\rightarrow$W & \cellcolor[HTML]{32CB00}MGDA$\rightarrow$W & \cellcolor[HTML]{C0C0C0}B$\rightarrow$MGDA & \cellcolor[HTML]{C0C0C0}B$\rightarrow$W & \cellcolor[HTML]{C0C0C0}MGDA$\rightarrow$W & \cellcolor[HTML]{C0C0C0}B$\rightarrow$MGDA & \cellcolor[HTML]{C0C0C0}B$\rightarrow$W & \cellcolor[HTML]{C0C0C0}MGDA$\rightarrow$W \\
DTLZ4& \cellcolor[HTML]{FE0000}B$\rightarrow$MGDA & \cellcolor[HTML]{C0C0C0}B$\rightarrow$W & \cellcolor[HTML]{32CB00}MGDA$\rightarrow$W & \cellcolor[HTML]{C0C0C0}B$\rightarrow$MGDA & \cellcolor[HTML]{C0C0C0}B$\rightarrow$W & \cellcolor[HTML]{32CB00}MGDA$\rightarrow$W & \cellcolor[HTML]{FE0000}B$\rightarrow$MGDA & \cellcolor[HTML]{C0C0C0}B$\rightarrow$W & \cellcolor[HTML]{32CB00}MGDA$\rightarrow$W & \cellcolor[HTML]{FE0000}B$\rightarrow$MGDA & \cellcolor[HTML]{C0C0C0}B$\rightarrow$W & \cellcolor[HTML]{32CB00}MGDA$\rightarrow$W \\
DTLZ5& \cellcolor[HTML]{32CB00}B$\rightarrow$MGDA & \cellcolor[HTML]{32CB00}B$\rightarrow$W & \cellcolor[HTML]{32CB00}MGDA$\rightarrow$W & \cellcolor[HTML]{32CB00}B$\rightarrow$MGDA & \cellcolor[HTML]{32CB00}B$\rightarrow$W & \cellcolor[HTML]{32CB00}MGDA$\rightarrow$W & \cellcolor[HTML]{32CB00}B$\rightarrow$MGDA & \cellcolor[HTML]{32CB00}B$\rightarrow$W & \cellcolor[HTML]{FE0000}MGDA$\rightarrow$W & \cellcolor[HTML]{FE0000}B$\rightarrow$MGDA & \cellcolor[HTML]{32CB00}B$\rightarrow$W & \cellcolor[HTML]{32CB00}MGDA$\rightarrow$W \\
DTLZ6& \cellcolor[HTML]{FE0000}B$\rightarrow$MGDA & \cellcolor[HTML]{FE0000}B$\rightarrow$W & \cellcolor[HTML]{32CB00}MGDA$\rightarrow$W & \cellcolor[HTML]{FE0000}B$\rightarrow$MGDA & \cellcolor[HTML]{FE0000}B$\rightarrow$W & \cellcolor[HTML]{FE0000}MGDA$\rightarrow$W & \cellcolor[HTML]{C0C0C0}B$\rightarrow$MGDA & \cellcolor[HTML]{32CB00}B$\rightarrow$W & \cellcolor[HTML]{32CB00}MGDA$\rightarrow$W & \cellcolor[HTML]{FE0000}B$\rightarrow$MGDA & \cellcolor[HTML]{FE0000}B$\rightarrow$W & \cellcolor[HTML]{32CB00}MGDA$\rightarrow$W \\
DTLZ7& \cellcolor[HTML]{FE0000}B$\rightarrow$MGDA & \cellcolor[HTML]{32CB00}B$\rightarrow$W & \cellcolor[HTML]{32CB00}MGDA$\rightarrow$W & \cellcolor[HTML]{FE0000}B$\rightarrow$MGDA & \cellcolor[HTML]{FE0000}B$\rightarrow$W & \cellcolor[HTML]{FE0000}MGDA$\rightarrow$W & \cellcolor[HTML]{C0C0C0}B$\rightarrow$MGDA & \cellcolor[HTML]{32CB00}B$\rightarrow$W & \cellcolor[HTML]{32CB00}MGDA$\rightarrow$W & \cellcolor[HTML]{FE0000}B$\rightarrow$MGDA & \cellcolor[HTML]{32CB00}B$\rightarrow$W & \cellcolor[HTML]{32CB00}MGDA$\rightarrow$W
\end{tabular}
}
    
    \caption{Pairwise significance map summarising the acceleration performance across all benchmark problems.
Within each method group the baseline (B) is compared to the catalyst versions and the catalysts to each other. Specifically the three columns from left to right correspond to Bonferroni-corrected ($\alpha = 0.05/4$) one-tailed Wilcoxon signed-rank tests for (i) B$\rightarrow$MGDA, (ii) B$\rightarrow$W, and (iii) MGDA$\rightarrow$W.
Green denotes a significant improvement of the challenger (right of the arrow) over the reference method (left of the arrow); 
Red indicates significant degradation. 
Grey marks the equivalence region, where either directional hypothesis is rejected, and the methods cannot be distinguished statistically.}
    \label{fig:traffic-light}
\end{figure}
%\todo{I think it would be better to use $M$ instead of $d$ or better remove $d$ in Figure 3: JF:DONE}

Across the seven problems and eight catalyst–baseline pairings, this results in 56 comparisons. Additionally, we show the 28 comparisons between the catalyst-pairs across each problem and acquisition function combination. In Figure~\ref{fig:traffic-light}, green cells indicate that acceleration significantly outperformed the corresponding baseline, red cells indicate degradation relative to the baseline, and grey cells denote no statistically significant difference. Of the 56 cases, 18 are green, 16 are red, and 22 are grey, excluding the comparisons between the catalyst variants MGDA and W. Thus, in approximately 32\% of cases acceleration leads to statistically significant improvements, while in a further 40\% of cases the accelerated and baseline methods behave equivalently. 
These results indicate that using a catalyst is generally beneficial, as only 28\% of cases result in degradation.

This motivates an examination of whether the quality of the surrogate model helps explain when acceleration is beneficial. To this end, Tables~\ref{tab:mad-median-group-1}--\ref{tab:mad-median-group-4} report, for each problem, method, and objective, the median of medians of the SLLs, together with its MAD. 

\begin{table*}[t!]
\centering
\scriptsize
\setlength{\tabcolsep}{4pt}
\begin{tabular}{lcccccccc}
\toprule
Method & \multicolumn{4}{c}{DTLZ1} & \multicolumn{4}{c}{DTLZ2} \\
 & $f_1$\,Med & $f_1$\,MAD & $f_2$\,Med & $f_2$\,MAD & $f_1$\,Med & $f_1$\,MAD & $f_2$\,Med & $f_2$\,MAD \\ \midrule
EHVI & -0.750 & 1.310 & -1.044 & 0.987 & -3.250 & 2.365 & -0.665 & 0.716 \\
MGDA-EHVI & -0.354 & 0.774 & -0.694 & 0.628 & -10.274 & 0.041 & -10.275 & 0.030 \\
W-EHVI & 0.209 & 1.123 & -0.649 & 0.260 & -10.039 & 0.051 & -10.029 & 0.050 \\
AugTch & 0.284 & 1.096 & -0.961 & 1.025 & -9.631 & 0.131 & -9.399 & 0.326 \\
MGDA-AugTch & -0.630 & 1.403 & -0.124 & 1.507 & -9.872 & 0.095 & -9.937 & 0.079 \\
W-AugTch & -0.809 & 0.436 & -1.306 & 1.239 & -10.037 & 0.082 & -10.063 & 0.059 \\
tMPoI & -0.137 & 0.446 & -0.900 & 0.847 & -4.788 & 0.487 & -3.849 & 0.507 \\
MGDA-tMPoI & -0.529 & 0.467 & 0.306 & 1.221 & -9.025 & 0.225 & -9.131 & 0.100 \\
W-tMPoI & -1.423 & 1.738 & -0.179 & 0.893 & -9.273 & 0.175 & -9.153 & 0.200 \\
SAF & -0.444 & 0.601 & -1.252 & 0.845 & -5.204 & 1.105 & -5.689 & 0.705 \\
MGDA-SAF & -1.140 & 1.131 & -0.214 & 0.507 & -10.228 & 0.066 & -10.285 & 0.065 \\
W-SAF & -0.750 & 0.967 & -0.627 & 1.450 & -10.038 & 0.070 & -10.107 & 0.053 \\
\bottomrule
\end{tabular}
\caption{Median-of-median-SLLs and MAD for objectives $f_1$ and $f_2$ for problems DTLZ1, DTLZ2.}
\label{tab:mad-median-group-1}
\end{table*}

\begin{table*}[t!]
\centering
\scriptsize
\setlength{\tabcolsep}{4pt}
\begin{tabular}{lcccccccc}
\toprule
Method & \multicolumn{4}{c}{DTLZ3} & \multicolumn{4}{c}{DTLZ4} \\
 & $f_1$\,Med & $f_1$\,MAD & $f_2$\,Med & $f_2$\,MAD & $f_1$\,Med & $f_1$\,MAD & $f_2$\,Med & $f_2$\,MAD \\ \midrule
EHVI & -0.374 & 2.783 & -0.551 & 1.936 & 1.154 & 4.193 & -1.361 & 3.166 \\
MGDA-EHVI & -0.902 & 3.939 & 0.136 & 2.355 & 9.810 & 11.732 & -6.392 & 2.441 \\
W-EHVI & -0.164 & 1.055 & -0.175 & 3.459 & -10.297 & 0.824 & -10.006 & 0.619 \\
AugTch & -0.582 & 1.375 & -1.150 & 3.249 & -4.454 & 2.190 & -3.720 & 1.829 \\
MGDA-AugTch & -0.592 & 0.567 & -0.703 & 0.973 & -5.729 & 2.977 & -9.590 & 0.608 \\
W-AugTch & -1.010 & 0.861 & -0.448 & 0.265 & -9.086 & 1.571 & -8.960 & 0.900 \\
tMPoI & -0.421 & 1.537 & -2.061 & 2.389 & -1.388 & 2.086 & -2.158 & 1.644 \\
MGDA-tMPoI & -0.426 & 1.851 & -0.721 & 1.168 & -3.370 & 3.492 & -10.186 & 2.453 \\
W-tMPoI & -0.242 & 1.178 & -1.134 & 1.250 & -9.352 & 1.017 & -9.596 & 0.817 \\
SAF & -0.057 & 0.257 & -0.346 & 0.742 & 2.906 & 3.723 & -1.311 & 3.427 \\
MGDA-SAF & -1.591 & 1.153 & -0.708 & 0.441 & 39.487 & 43.987 & -6.003 & 1.895 \\
W-SAF & -1.164 & 0.680 & -0.379 & 1.338 & -9.271 & 1.347 & -8.912 & 1.367 \\
\bottomrule
\end{tabular}
\caption{%Median-of-median-SLLs and MAD for objectives $f_1$ and $f_2$ for 
Details as for Table \ref{tab:mad-median-group-1}, DTLZ3 and DTLZ4 problems.}
\label{tab:mad-median-group-2}
\end{table*}

\begin{table*}[t!]
\centering
\scriptsize
\setlength{\tabcolsep}{4pt}
\begin{tabular}{lcccccccc}
\toprule
Method & \multicolumn{4}{c}{DTLZ5} & \multicolumn{4}{c}{DTLZ6} \\
 & $f_1$\,Med & $f_1$\,MAD & $f_2$\,Med & $f_2$\,MAD & $f_1$\,Med & $f_1$\,MAD & $f_2$\,Med & $f_2$\,MAD \\ \midrule
EHVI & -0.967 & 1.116 & -2.815 & 2.319 & -2.633 & 1.135 & -0.653 & 1.840 \\
MGDA-EHVI & -10.301 & 0.015 & -10.308 & 0.041 & -2.000 & 0.455 & -1.361 & 1.546 \\
W-EHVI & -10.074 & 0.023 & -10.080 & 0.063 & -2.248 & 1.161 & -2.422 & 1.225 \\
AugTch & -9.453 & 0.379 & -9.488 & 0.304 & -3.065 & 3.009 & 11.589 & 18.372 \\
MGDA-AugTch & -9.778 & 0.069 & -9.869 & 0.038 & -1.482 & 0.603 & -1.719 & 1.180 \\
W-AugTch & -10.058 & 0.104 & -10.007 & 0.086 & -2.702 & 0.550 & -2.237 & 1.212 \\
tMPoI & -3.985 & 0.529 & -4.327 & 0.771 & 0.070 & 0.341 & -0.895 & 2.827 \\
MGDA-tMPoI & -9.002 & 0.264 & -9.024 & 0.165 & -0.691 & 2.859 & -0.053 & 8.415 \\
W-tMPoI & -9.518 & 0.158 & -9.337 & 0.257 & 0.212 & 4.208 & -0.620 & 1.021 \\
SAF & -5.196 & 0.427 & -4.724 & 1.095 & -2.629 & 2.398 & -0.606 & 0.529 \\
MGDA-SAF & -10.236 & 0.017 & -10.245 & 0.048 & -3.157 & 0.777 & -2.291 & 1.127 \\
W-SAF & -10.024 & 0.010 & -10.039 & 0.042 & -2.684 & 0.632 & -2.203 & 0.591 \\
\bottomrule
\end{tabular}
\caption{%Median-of-median-SLLs and MAD for objectives $f_1$ and $f_2$ for problems 
Details as for Table \ref{tab:mad-median-group-1}, DTLZ5 and DTLZ6 problems.}
\label{tab:mad-median-group-3}
\end{table*}

\begin{table*}[t!]
\centering
\scriptsize
\setlength{\tabcolsep}{4pt}
\begin{tabular}{lcccc}
\toprule
Method & \multicolumn{4}{c}{DTLZ7} \\
 & $f_1$\,Med & $f_1$\,MAD & $f_2$\,Med & $f_2$\,MAD \\ \midrule
EHVI & -11.490 & 0.502 & -4.658 & 4.289 \\
MGDA-EHVI & -9.926 & 0.502 & 7.202 & 7.369 \\
W-EHVI & -12.122 & 0.096 & -9.496 & 0.163 \\
AugTch & -12.630 & 0.066 & -11.375 & 0.111 \\
MGDA-AugTch & -13.089 & 0.111 & -9.349 & 0.504 \\
W-AugTch & -10.613 & 0.463 & -0.501 & 1.967 \\
tMPoI & -12.137 & 0.383 & 0.147 & 4.938 \\
MGDA-tMPoI & -12.915 & 0.610 & -8.348 & 2.908 \\
W-tMPoI & -12.392 & 0.116 & -10.181 & 0.036 \\
SAF & -12.132 & 0.165 & 5.772 & 4.698 \\
MGDA-SAF & -11.211 & 0.342 & -0.260 & 3.015 \\
W-SAF & -12.606 & 0.061 & -9.495 & 0.092 \\
\bottomrule
\end{tabular}
\caption{%Median-of-median-SLLs and MAD for objectives $f_1$ and $f_2$ for 
Details as for Table \ref{tab:mad-median-group-1}, DTLZ7 problem.}
\label{tab:mad-median-group-4}
\end{table*}

\paragraph{DTLZ1 and DTLZ3.}
For DTLZ1, Figure~\ref{fig:traffic-light} shows all grey cells, indicating that acceleration does not substantially affect performance. This aligns with the surrogate quality metrics: for both objectives, median SLL values are only modestly negative, while MAD values are sufficiently large to span both negative and positive values. This indicates inconsistent and only moderately informative surrogate models. A similar pattern is observed for DTLZ3, where surrogate quality is comparable, with median SLLs near zero and large MADs frequently extending into positive values. Consequently, most comparisons remain grey, suggesting that poor or highly variable surrogate quality limits the effectiveness of acceleration.

\paragraph{DTLZ2.}
In contrast, DTLZ2 shows widespread green cells in Figure~\ref{fig:traffic-light}, indicating a consistent improvement under acceleration. This coincides with excellent surrogate quality: median SLL values are strongly negative (often approaching $-10$), MAD values are small, and both objectives are predicted reliably across runs. These results indicate that high-quality surrogates can effectively support accelerated acquisition by accurately identifying Pareto-relevant regions.

\paragraph{DTLZ4.}
Results for DTLZ4 are mixed. In several cases, most notably SAF combined with MGDA acceleration, the quality of the surrogate model for the first objective is extremely poor, with median SLL values exceeding $+30$. Such values indicate models that are effectively uninformative. In these cases, acceleration is detrimental, as reflected by the red cells in Figure~\ref{fig:traffic-light}. More broadly, DTLZ4 is characterised by large MADs for the first objective, leading to unstable surrogate performance across runs and limiting the reliability of acceleration.

\paragraph{DTLZ5.}
For DTLZ5, the quality of the surrogate model is consistently strong in both objectives. Median SLL values are typically around $-10$, with small MADs. Correspondingly, acceleration improves performance in most cases, as shown in Figure~\ref{fig:traffic-light}. The only notable exception occurs when SAF is combined with MGDA, where the baseline already performs competitively. In general, robust and stable surrogate models enable effective acceleration for this problem.

\paragraph{DTLZ6.}
In DTLZ6, surrogate quality is moderate but inconsistent. Median SLL values tend to cluster around $-2$, while MADs are often of comparable magnitude. In some cases, particularly for the second objective, median SLL values become positive. This instability aligns with the predominance of red cells in Figure~\ref{fig:traffic-light}, indicating that acceleration often degrades performance. Notably, only tMPoI combined with predefined weights yields consistent improvement, suggesting that careful choice of acceleration strategy can partially mitigate imperfect surrogate quality.

\paragraph{DTLZ7.}
Finally, DTLZ7 presents an asymmetric situation. The first objective is modelled extremely well, with strongly negative median SLL values and small MADs. In contrast, the second objective model often exhibits larger MADs and occasional positive median SLLs. As a result, acceleration yields modest improvements relative to DTLZ6, where both objective surrogates were poor—particularly for the weighted variants of EHVI, tMPoI, and SAF.

\paragraph{Stationarity analysis.}
To further interpret these results, we examine the stationarity assumption that underpins the stationary kernels used in the surrogate models. Although all DTLZ problems are continuous and differentiable, stationarity additionally requires that local perturbations in the decision space induce similar variations in the objective space, irrespective of location.

\begin{figure}[t!]
    \centering
    \includegraphics[width=\linewidth, trim={10 22 10 5mm}, clip=true]{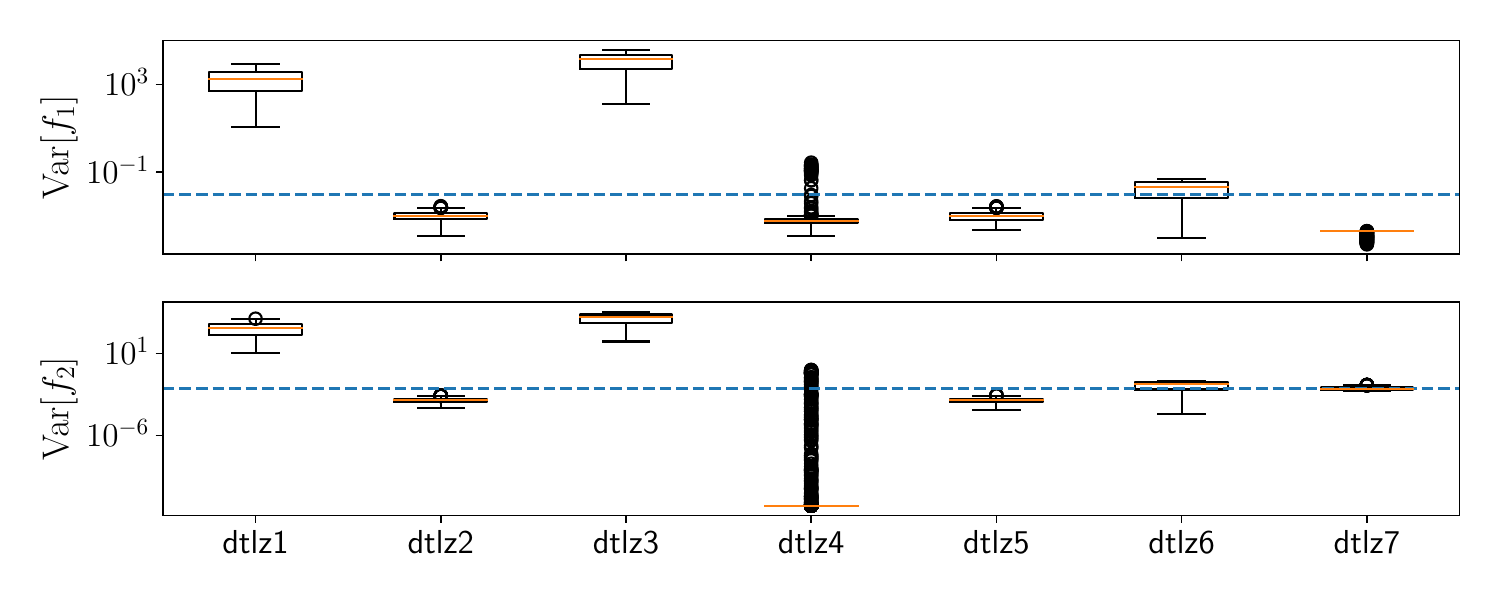}
    \caption{
Local variance of objectives $f_1$ (top) and $f_2$ (bottom), computed from
$1000$ randomly placed local neighbourhoods in a $10$‑dimensional decision
space and their corresponding objective values. Each neighbourhood spans
$5\%$ of the domain per dimension and contains $1000$ Latin–hypercube samples.
The dashed horizontal line indicates variance $=10^{-2}$, which may be regarded
as a small variance threshold suggestive of approximate stationarity over the
domain. DTLZ2 and DTLZ5 exhibit the most stationary behaviour, while DTLZ1 and
DTLZ3 show clear non‑stationarity. For DTLZ4, several outliers exceed the
threshold, indicating locally sharp variations despite otherwise smooth
behaviour. In DTLZ7, the first objective appears largely stationary, whereas
the second objective lies closer to the boundary of the threshold. DTLZ6
exhibits regions with pronounced jumps in variance relative to the rest of
the domain.
}
    \label{fig:dtlz-stationarity}
\end{figure}

To test this assumption, we sample 1000 random ten-dimensional hyper-boxes in the decision space, each spanning 5\% of the domain per dimension and populated with 1000 Latin-hypercube samples. For each box, we compute the local output variance for both objectives. The resulting distributions, shown in Figure~\ref{fig:dtlz-stationarity}, reveal pronounced differences across problems.

DTLZ2 and DTLZ5 exhibit uniformly low local variances across the domain, remaining below a variance threshold of $10^{-2}$ and indicating strong stationarity, which we attribute as a key reason for the superior performance of the catalytic framework. In contrast, DTLZ1 and DTLZ3 show substantial variation in local variance, confirming pronounced non-stationarity, and therefore, the catalytic framework offers limited benefit. DTLZ4 displays frequent outliers beyond the threshold—particularly for the first objective—consistent with its mixed performance under acceleration. DTLZ6 similarly exhibits intermittent large local variances for both objectives, resulting in poor performance of acceleration strategies. Finally, DTLZ7 shows near-stationary behaviour in the first objective but borderline or non-stationary behaviour in the second, explaining its intermediate performance.

% Overall, these results demonstrate that the stationarity of the problem and stability of the surrogate model are key determinants of whether acceleration improves performance. Even for %fully 
% smooth benchmark problems, violations of stationarity can undermine the effectiveness of surrogate-driven acceleration strategies.
Overall, when a problem is stationary and consistent with GP kernel assumptions, surrogate-driven acceleration is almost always beneficial: W is always superior to the baseline, while MGDA is outperformed by or equivalent to the baseline only with SAF (in DTLZ5 and DTLZ2 respectively). It should be noted that only two of the seven problems tested exhibit stationary behaviour.

Finally, we note an important distinction between the two catalytic approaches, MGDA and W. Across the 28 pairwise comparisons between these variants, only four cases favour MGDA, while seven cases result in statistical equivalence (see Figure~\ref{fig:traffic-light}). In the remaining comparisons, W is the superior performer, indicating a clear overall advantage.

From a conceptual perspective, MGDA can be interpreted as allowing an effectively infinite number of descent directions, with an optimal direction identified locally for each solution. In contrast, the W variant operates on a coarse, finite set of predefined directions. This reduced granularity may be beneficial in the context of expensive optimisation problems, because of the limited budget on the number of function evaluations, where focusing the search on a limited subset of promising regions can be more effective than attempting to finely approximate the entire global Pareto set.

The experiments are limited to the DTLZ benchmark suite with two objectives and ten decision variables. The framework should nonetheless scale well to higher-dimensional settings, as predictive gradients are analytically derived from independent GP models and the MGDA solver operates on the Gram matrix of objective gradients, keeping computational overhead modest, even for tens of objectives. Whether the stationarity observations generalise to higher-dimensional problems remains an open question for future work.

\section{Conclusion}
\label{sec:concl}

This paper introduced a predictive‑gradient catalytic framework for accelerating multi-objective Bayesian optimisation under limited evaluation budgets. By exploiting surrogate predictive gradients as auxiliary signals, the proposed approach augments standard Pareto-compliant acquisition functions with local stationarity information, without altering their underlying structure. Two catalyst instantiations were investigated: an adaptive MGDA-based strategy grounded in Pareto stationarity theory, and a predefined-weight variant that enables robust, focused exploration when budgets are tight. Experiments on the DTLZ benchmark suite show that predictive‑gradient catalysis can deliver significant acceleration when surrogate models are accurate, particularly for stationary problems. These results highlight surrogate-derived gradients as a promising mechanism for improving efficiency in expensive multi-objective optimisation and motivate extensions to constrained and noisy settings. Future works include testing on high dimensional problems, both in objective and decision spaces and adapting the weights (e.g. taking a similar approach to \cite{10.1007/978-981-96-3538-2_17}) for the acquisition function and predictive gradients in Equation~\ref{eq:tch_inside}.

\begin{credits}
%\subsubsection{Licence} For the purpose of open access, the author has applied a Creative Commons Attribution (CC BY) licence to any Author Accepted Manuscript version arising from this submission.
\subsubsection{\discintname}
The authors have no competing interests to declare that are
relevant to the content of this article. 
\subsubsection{\ackname} 
We thank the organisers of the Dagstuhl Seminar \emph{Multi-objective Optimization: Theory, Methods and Applications}
(\href{https://www.dagstuhl.de/en/seminars/seminar-calendar/seminar-details/23361}{Seminar~23361})
for creating a stimulating environment that influenced this work. We are particularly grateful to Kaisa Miettinen for insightful discussions and feedback, and to Tea Tu\u{s}ar, Dimo Brockhoff, and Pascal Kerschke for valuable conversations. Rahat acknowledges support from the Engineering and Physical Sciences Research Council [grant number EP/W01226X/1].
% We would like to thank the organisers of the Dagstuhl Seminar
% \emph{Multi-objective Optimization: Theory, Methods and Applications}
% (Seminar~23361) for providing a stimulating and collaborative environment that greatly benefited this work. Many of the ideas presented here were shaped through discussions during and following the seminar. We are particularly indebted to Kaisa Miettinen for insightful conversations and valuable feedback. We also thank Tea Tu\u{s}ar, Dimo Brockhoff, and Pascal Kerschke for engaging and thought-provoking discussions that helped refine our perspective.

% Any to add? Dagstuhl, Kaisa, Tea Tusar, Dimo, Pascal, etc. Any one else? 

%A bold run-in heading in small font size at the end of the paper is used for general acknowledgments, for example: This study was funded by X (grant number Y).

% I think this declaration can also me made in the submission system which means the disclosure of interests section can be omitted in the text.

%\subsubsection{Declaration of Generative AI Use.}
%The authors used a generative AI tool to support language editing, LaTeX formatting, and refinement of the presentation. The tool did not generate scientific results or influence the experimental design, analysis, or conclusions.
\end{credits}

\bibliographystyle{splncs04}
\bibliography{references}

\end{document}